\documentclass[11pt,a4paper]{article}
\usepackage[hyperref]{emnlp2020}
\usepackage{times}
\usepackage{latexsym}
\usepackage{multirow}

\usepackage{microtype}

\aclfinalcopy 

\title{Lessons Learned from Applying off-the-shelf BERT: There is no Silver Bullet}

\author{Victor Makarenkov \\
  \texttt{vitiokm@gmail.com} \\\And
  Lior Rokach \\
  Ben-Gurion University of the Negev \\
  \texttt{liorrk@post.bgu.ac.il} \\}

\date{}

\begin{document}
\maketitle
\begin{abstract}
One of the challenges in  the NLP field is training  large  classification  models, a task that is both difficult and tedious. It is even harder when GPU hardware is unavailable. The increased availability of pre-trained and off-the-shelf word embeddings, models, and modules aim at easing the process of training large models and achieving a competitive performance. 

We explore the use of off-the-shelf BERT models and share the results of our experiments and compare their results to those of LSTM networks and more simple baselines. We show that the complexity and computational cost of BERT is not a guarantee for enhanced predictive performance in the classification tasks at hand. 
\end{abstract}

\section{Introduction}

Deep learning methods have revolutionized the NLP field in the past ten years. Although LSTM networks \cite{LSTM} have been around for over two decades, the NLP community only learned how to train and use them effectively in the past ten years. \citet{DBLP:journals/corr/ChoMGBSB14} introduced a new sequence-to-sequence method, boosting the field of neural machine translation significantly \cite{seq2seq}. The same year \citet{bahdanau2014neural} presented the attention mechanism aimed at focusing on specific words within the prefix, in order to make the most accurate prediction of the next word while mapping one sequence to another. During the same period new text representation methods were adapted, complementing the following representation methods: bag-of-words (BOW), tf-idf, and one-hot vectors with dense representations, such as the very prominent word2vec \cite{mikolov2013efficient} and Glove \cite{Pennington14glove:global} embeddings, which served as the go-to methods in many works \cite{10.1145/3159652.3159733, blumenthal-graf-2019-utilizing}.

\citet{devlin2018bert} introduced a pre-trained transformer \cite{vaswani2017attention} based on the attention mechanism without any recurrent connections. BERT provided another advancement in the field of pre-trained text representations, showing enhanced performance on various NLP tasks \cite{devlin2018bert, goldberg2019assessing}. 

Many research directions were shaped by  pre-trained word embeddings and representations with several software toolkits available for training deep neural networks. While the Keras \cite{chollet2015keras} toolkit was widely used for text classification \cite{10.1145/3159652.3159733} with padding, the DyNet \cite{dynet} and PyTorch \cite{NEURIPS2019_9015} toolkits excelled at tasks in which a dynamic computation graph of the recurrent networks was exploited to achieve better predictive performance with sentences of varying length \cite{aharoni2016improving, ziser2018pivot}.

An important advancement in the dense representation area occurred with the introduction of TensorFlow \cite{tensorflow2015-whitepaper} Hub in 2018. According to Google\footnote{https://www.tensorflow.org/hub}: "TensorFlow Hub is a library for the publication, discovery, and consumption of reusable parts of machine learning models". Google introduced the term \textit{module} which refers to a model's component that can be easily accessible from code via TensorFlow Hub and integrated in the desired model for a downstream task. Today, many pre-trained off-the-shelf models and modules are available from TensorFlow Hub, where they are accessible with the widely-familiar and user friendly Keras look-and-feel. TensorFlow Hub offers multiple pre-trained BERT modules. The availability of these modules actually provides a range of possible representation options for the input text as an interchangeable components of a model. 

Pre-trained BERT models and other TensorFlow modules are usually trained and tested over large  open datasets. In contrast, we focused on applying these models on small proprietary datasets.
In this paper, we share our experience using these promising and accessible text representation methods and tools. With these methods and tools, we aimed to outperform advanced baselines based on proprietary word embeddings and LSTM networks in two different tasks. We conducted a series of experiments with modules available from Tensorflow Hub and pre-trained BERT modules which we fine-tuned for the downstream task. Surprisingly, the results we achieved were not as good as the baselines. In Section 2, we discuss the experiments on the \textit{proper word choice} task. In Section 3, we present the experiments on the \textit{political perspective detection} task. We openly share the datasets which we used in the course of this research\footnote{\url{https://github.com/vicmak/News-Bias-Detection}}.

\section{Proper Word Choice}

\begin{table*}[]
\centering
\caption{MRR values of BiLSTM and N-Gram models as opposed to MRR values achieved by off-the-shelf and fine tuned BERT models}
\label{table:writing:results}
\begin{tabular}{llc}
\hline
 & \textbf{Model} & \textbf{MRR} \\ \hline
\multirow{3}{*}{\textbf{Baselines}} & BiLSTM - domain specific & 0.41 \\
 & BiLSTM - general purpose (COCA) & 0.33 \\
 & N-gram - domain specific & 0.34 \\ \hline
\multirow{3}{*}{\textbf{BERT-based}} & bert-base-uncased & 0.29 \\
 & bert-large-uncased & 0.30 \\
 & fine-tuned bert-base-uncased & 0.31 \\ \hline
\end{tabular}
\end{table*}

\subsection{Related Tasks}
Over the years, several methods proposed by researchers in the NLP community attempted to solve the task of changing a word withing a given text. Among the most notable are: 1) grammatical error correction  \cite{DBLP:conf/acl/RozovskayaR16, ng-EtAl:2014:W14-17}, 2) lexical substitution \cite{mccarthy2009english} , 3) choosing the most typical word in context \cite{edmonds1997choosing}, and 4) cloze \cite{taylor1953cloze}, which is referred to as masked language modeling (MLM) by \citet{devlin2018bert}.

\subsection{The Proper Word Choice Task}

Given a sentence:
\begin{equation}
    s=<w_1, w_2,...,target,...,w_n>
\end{equation}{}
 where the $target$ is explicitly specified, the task is to provide a sorted list of the $k$ most appropriate words to replace the $target$ in $s$ based on the sentential context of the $target$ in $s$. 

For example, the following proper word choices (in green) can be made instead of the original (in red) words.

\begin{itemize}
    \item My wife thinks that I am a \textcolor{red}{handsome} $\rightarrow$ \textcolor{green}{beautiful} guy.
    \item I always drink a \textcolor{red}{powerful} $\rightarrow$ \textcolor{green}{strong} tea.
    \item The results clearly \textcolor{red}{indicate} $\rightarrow$ \textcolor{green}{show} that our method significantly outperforms the current state-of-the-art.
\end{itemize}{}

\subsection{The Dataset}

We used a corpus of 30,000 academic articles \cite{makarenkov2019choosing} collected from 60 top ranked $A$ and $A^*$ ACM conferences to train from scratch a bidirectional LSTM model as a baseline. We use the same corpus to \textit{fine tune} the off-the-shelf pre-trained BERT models. This corpus consists of  of roughly 40 million tokens in 2.78 million sentences. 

\subsection{Evaluation Settings}

For the evaluation we use a test-set\footnote{Available here: \url{https://github.com/vicmak/Exploiting-BiLSTM-for-Proper-Word-Choice}} of 176 sentences \cite{makarenkov2019choosing} which include a specified word that was replaced during the process of professional editing by native English editors. We use the mean reciprocal rank (MRR) metric to test the model effectiveness. We use the HuggingFace \cite{wolf2019huggingfaces} toolkit to process the data and fine-tune the pre-trained BERT model. We used the Google's Colab environment with GPU acceleration to train the model.

We started our experiments by initializing our model with an off-the-shelf pre-trained BERT model. We then applied the MLM functionality to the proper word choice. 

In the first experiment we used the \texttt{bert-base-uncased} model. In the second experiment we used the \texttt{bert-large-uncased} model. In the third experiment we initialized our model with \texttt{bert-base-uncased} and then fine-tuned the model with the scientific corpus for the downstream task of MLM.

\subsection{Results}

We consider an advanced baseline that uses a bidirectional LSTM tagger \cite{DBLP:conf/conll/MelamudGD16}, and was trained from scratch on the domain specific scientific corpus.
The results are presented in Table \ref{table:writing:results}. The three bottom rows in Table \ref{table:writing:results} show the BERT-based results, while the three upper rows contain the results of the baselines. Surprisingly, the results obtained by the BERT-based models failed to outperform the BiLSTM models as well as the classical n-gram model with Kneser-Ney \cite{NEY19941} smoothing.

A bidirectional LSTM model trained from scratch on the same domain-specific corpus as the fine-tuned BERT model outperformed it by 25\% in terms of the MRR. A bidirectional LSTM model trained from scratch on the general-purpose COCA \cite{COCA} corpus achieved superior performance as well.  

\section{Political Perspective Detection}

\begin{table*}[]
\centering
\caption{The results of political perspective detection with different models.}
\label{table:news}
\begin{tabular}{llccl}
\hline
 & \textbf{Model} & \multicolumn{1}{l}{\textbf{\begin{tabular}[c]{@{}l@{}}Israeli perspective \\ recall\end{tabular}}} & \multicolumn{1}{l}{\textbf{\begin{tabular}[c]{@{}l@{}}Palestinian perspective \\ recall\end{tabular}}} & \textbf{AUC} \\ \hline
\multirow{3}{*}{\textbf{Baselines}} & SVM & 0.833 & 0.586 & 0.710 \\
 & Logistic Regression & 0.963 & 0.845 & 0.964 \\
 & LSTM & 0.966 & 0.879 & 0.966 \\ \hline
\multirow{2}{*}{\textbf{Off-the-shelf models}} & \begin{tabular}[c]{@{}l@{}}BERT pre-trained\\ fine-tuned\end{tabular} & 0.778 & 0.638 & 0.797 \\
 & \begin{tabular}[c]{@{}l@{}}Swivel news \\ matrix factorization\end{tabular} & 0.740 & 0.766 & 0.840 \\ \hline
\end{tabular}
\end{table*}

\subsection{The Task}

Many European and American media companies provide presumably neutral online news articles in their coverage of the Israeli-Palestinian conflict, one of the most longstanding, debated, and emotionally charged conflicts in the world. This task aims on identifying the political perspective of an article in the European or American press. The task is seen as a binary text classification task, determining whether the article reflects the Israeli or Palestinian perspective of the conflict. 

\subsection{The Dataset}

To train the model we used a dataset of 25,000 articles \cite{makarenkov2019implicit} which are covering the conflict from both the \textit{Israeli} and \textit{Arab} online media.
The articles from the Israeli media were given an Israeli-perspective label, and the articles from the Arab media sources were given a Palestinian-perspective label. 

For example, the following Al Jazeera article\footnote{https://www.aljazeera.com/news/middleeast/2014/01/rights-groups-condemn-israel-raids-homes-2014122582145919.html}, which was given a Palestinian perspective label was used in training: "\textit{Israel raids on West Bank homes condemned}". A Times of Israel article\footnote{https://www.timesofisrael.com/top-us-official-slams-israeli-criticism-of-kerry/} entitled "\textit{Top US official pans Israeli criticism of Kerry}", which was given an Israeli perspective label,  was used in training.

\subsection{Evaluation Settings}

We used a test-set that contains 113 articles from the British and from American press. The articles were manually annotated by different annotators who agreed on the label given. Fifty four articles in the test set are labeled with an Israeli perspective and fifty nine with a Palestinian perspective. All of the articles in the test set originate from presumably neutral \textit{American} and \textit{British} media sources that were not used in the training.

For example, the following BBC-news article\footnote{https://www.bbc.com/news/world-middle-east-33128955} was agreed by the annotators to have a Palestinian perspective: "\textit{Gaza war actions lawful, report says}". A Fox-news article\footnote{https://www.foxnews.com/opinion/tel-aviv-attacks-israelis-want-peace-but-need-a-peace-partner}, entitled "\textit{Tel Aviv attacks: Israelis want peace but need a peace partner}" was consensualy annotated as having an Israeli perspective.

We conducted a series of experiments with pre-trained off-the-shelf models which were downloaded as modules and used with Keras wrapping and TensorFlow Hub \cite{tensorflow2015-whitepaper}. We used one model as is, we fine-tuned the second one for the downstream task:
\begin{itemize}
    \item A general-purpose BERT  model\footnote{\url{https://tfhub.dev/tensorflow/bert_en_uncased_L-12_H-768_A-12/2}} trained on the English Wikipedia and Books corpus. This model contain 12 transformers blocks, 12 attention heads, and a hidden size of 768. We fine-tuned this model with the corpus of 25,000 articles used in the training phase.
    \item An off-the-shelf news-domain model based on Swivel co-occurrence matrix factorization\cite{DBLP:journals/corr/ShazeerDEW16}. The model was trained on English Google News 130GN corpus. Since this model was trained on news corpus, we aimed to explore the possibility of transfer learning in similar domains.
    
\end{itemize}{}

\subsection{Results}

The results for applying the fine-tuned and off-the-shelf models are presented in Table \ref{table:news}. The top three rows contain the results of the baselines, and the bottom two rows present the best results we achieved in a set of experiments performed using pre-trained models and different hyperparameters. All of the code and the test-set is available here\footnote{Anonymized link to GitHub repository}.

Both off-the-shelf models, even a fine-tuned BERT model, were outperformed by the baselines. Not only was the performance inferior to an LSTM model that also captures long term linguistic regularities; the performance of the BERT model was also outperformed by a classic SVM classifier that used tf-idf bag-of-words text representation in the Israeli perspective recall metric.

\section{Limitations}

In the proper word choice task the main limitation of the experiments line stems from the way tokenization is performed by BERT. BERT adopts the WordPiece \cite{wu2016googles} tokenization procedure in which the vocabulary is constructed incrementally starting at the character level, with respect to the corpus. As a result, the vocabulary that was used in the BERT models was different than the baseline models, which used the NLTK \cite{bird06nltk} library to tokenize the text during training; thus, the probability distribution was computed over a different lexicon in the task of choosing the proper word. One may claim that because of this the experiments are not directly comparable.
The difference in the tokenization procedure does not affect the \textit{comparability} of political perspective detection, since the downstream task is text classification and that's exactly what we sought to exploit in order to achieve an enhanced performance.

The second major limitation is that in both tasks the test sets are relatively small. There are 176 sentences in the test set for the task of proper word choice and  112 articles in the test set for political perspective identification task. 

\section{Discussion}

We share the results of our experimentation with two different NLP tasks. In both cases we experimented with small proprietary datasets from domains that suffer from a serious lack of labeled data. In these experiments we used the very promising and prominent BERT method and off-the-shelf TensorFlow Hub modules with the aim of outperforming several baselines on the tasks of proper word choice and political perspective identification. We used both pre-trained off-the-shelf and fine-tuned proprietary models. We failed to outdo the earlier folklore baselines as well as an advanced LSTM-based baselines, with a straightforward and systematic way of applying BERT.

Over 30 years ago, \citet{brooks1987no} argued that the software development process is hard at its very essence. They could not envision an advanced programming language capable of solving the complexity of performing high-quality software development projects on time. Analogously, a more user-friendly framework for pre-trained models can't guarantee excellent predictive performance. Training high performing models is essentially difficult. It requires deep understanding of the task data processing expertise. Fine-tuning a pre-trained model might be a good starting point, but the developer will still be required to delve deeply into a model's details in order to excel at predictive performance.


\bibliography{anthology,emnlp2020}
\bibliographystyle{acl_natbib}

\end{document}